\pdfoutput=1

\documentclass[11pt]{article}

\usepackage{acl}

\usepackage{times}
\usepackage{latexsym}

\usepackage[T1]{fontenc}

\usepackage[utf8]{inputenc}

\usepackage{microtype}

\usepackage{amsmath} 
\usepackage{amsfonts} 
\usepackage{graphicx} 
\usepackage{multirow,tabularx} 
\usepackage{booktabs} 
\usepackage{makecell} 
\usepackage{multirow} 
\usepackage{tabularx} 

\title{Improving Multi-Document Summarization through Referenced Flexible Extraction with Credit-Awareness}

\author{
  Yun-Zhu Song \and Yi-Syuan Chen \and Hong-Han Shuai \\
  National Yang Ming Chiao Tung University, Taiwan \\
  \texttt{\{yunzhusong.eed07g,yschen.eed09g,hhshuai\}@nctu.edu.tw} \\
}
\begin{document}
\maketitle
\newcommand\abbrev{REFLECT}

\begin{abstract}
A notable challenge in Multi-Document Summarization (MDS) is the extremely-long length of the input. In this paper, we present an extract-then-abstract Transformer framework to overcome the problem. Specifically, we leverage pre-trained language models to construct a hierarchical extractor for salient sentence selection across documents and an abstractor for rewriting the selected contents as summaries. However, learning such a framework is challenging since the optimal contents for the abstractor are generally unknown. Previous works typically create \textit{pseudo extraction oracle} to enable the supervised learning for both the extractor and the abstractor. Nevertheless, we argue that the performance of such methods could be restricted due to the insufficient information for prediction and inconsistent objectives between training and testing. To this end, we propose a loss weighting mechanism that makes the model aware of the unequal importance for the sentences not in the pseudo extraction oracle, and leverage the fine-tuned abstractor to generate summary references as auxiliary signals for learning the extractor. Moreover, we propose a reinforcement learning method that can efficiently apply to the extractor for harmonizing the optimization between training and testing. Experiment results show that our framework substantially outperforms strong baselines with comparable model sizes and achieves the best results on the Multi-News, Multi-XScience, and WikiCatSum corpora.\footnote{The implementation code and trained models are available at https://github.com/yunzhusong/NAACL2022-REFLECT}
\end{abstract}

\section{Introduction}
Neural multi-document summarization has drawn a lot of attention due to the wide applications, e.g., Wikipedia generation~\cite{liu2018generating}, news digest~\cite{fabbri-etal-2019-multi}, or related-work section generation~\cite{lu-etal-2020-multi-xscience}. Through concatenating the document clusters, it is applicable to adopt the approaches from single-document summarization~\cite{pmlr-v119-zhang20ae, fabbri-etal-2019-multi} under multi-document settings. However, the input length from multiple documents is typically long, which might be computationally infeasible for Transformer-based models~\cite{NIPS2017_3f5ee243}. One possible solution is to set a fixed length and truncate the input sequence. However, the truncation leads to information loss and performance drop. Moreover, even if the model can take such a long sequence as input, the attentions could be dispersed over long sequences~\cite{yang-etal-2018-modeling} and further degrade the performance~\cite{jin-etal-2020-multi,liu-lapata-2019-hierarchical,cohan-etal-2018-discourse}.

To tackle the long sequence problem, another possible solution is to leverage the extract-then-abstract architecture in single document summarization~\cite{pilault-etal-2020-extractive,chen-bansal-2018-fast, gehrmann-etal-2018-bottom}. In such an architecture, an extractor first selects salient contents from the documents, and an abstractor is applied to rewrite the selected contents to coherent summaries. However, several issues arise for directly applying the procedure in MDS. 
1) \textit{Long-sequence problem for the extractor}. Although the input length for the abstractor can be controlled through the content selection and thus enables the usage of Transformer models~\cite{pilault-etal-2020-extractive}, it is still inevitable for the extractor to process the complete input documents. Previous methods mainly use LSTM-based extractors~\cite{pilault-etal-2020-extractive,chen-bansal-2018-fast, gehrmann-etal-2018-bottom}. However, this fails to leverage knowledge of pre-trained language models.
2) \textit{Suboptimal pseudo oracles}. Most summarization corpora do not contain the oracle for extraction. As an alternative, previous works typically generate \textit{pseudo extraction oracle}, or simply called \textit{pseudo oracle}, through a greedy process~\cite{liu-lapata-2019-text, chen-bansal-2018-fast}. Specifically, for each iteration in the process, candidate sentences are individually concatenated with previously-selected sentences and compute the ROUGE~\cite{lin-2004-rouge} scores with the human-written summaries. The top-scored sentence is iteratively selected until no sentence could further improve the ROUGE score. In practice, there are different ways for scoring. For examples, \citet{liu-lapata-2019-text} use the average of ROUGE-1 and ROUGE-2 F1 scores while \citet{chen-bansal-2018-fast} use ROUGE-L recall. The variants of design could cause the extractor to behave differently. In our study, we also found that using the ROUGE precision metric leads to much less extraction than using the recall metric. This implies that the pseudo oracles are suboptimal, and learning the extractor fully relies on the pseudo oracles could restrict the performance. How to alleviate the negative effects of pseudo oracles remains open.
3) \textit{Insufficient information for the extractor}. Even if the pseudo oracles are good enough to train the abstractor well, learning a precise extractor is still challenging. The problem lies in that a pseudo oracle is derived from a specific summary. However, there are potentially multiple valid summaries given the documents. To select salient sentences, the extractor is required to implicitly infer the underlying summary used for oracle construction, which is difficult due to the lack of evidence.
4) \textit{Inconsistent objectives for the extractor}. With pseudo oracles, the extractor is learned to select a set of sentences that has high lexical similarity to the summaries without redundancy. However, the goal of the extractor in test-time is to provide inputs for the abstractor that non-overlapping lexical may still be valuable. In other words, the objective for the extractor in training is inconsistent with the one in testing.

To address these issues, we propose the \textbf{RE}ferenced \textbf{FLE}xible Extraction with \textbf{C}redi\textbf{T}-Awareness (REFLECT) for MDS. For the first problem, we propose a Transformer-based hierarchical extractor that contains the token- and sentence-level feature encoders. Both the encoders are initialized with pre-trained language models to utilize the pretext knowledge.
For the second problem, we propose Pseudo Oracle Relaxation (POR) to render the model aware of the unequal importance for the non-oracle sentences. This mechanism encourages the model to emphasize the precision for critical sentences with either high or low lexical similarity to the summaries, and avoids the confusion arising from the different labels for similar sentences.
For the third problem, we propose Summary Referencing (SR) to leverage the fine-tuned abstractor for providing additional learning signals to evidence extraction prediction. The summary reference serves as an approximation for the human-written summary while being able to generalize for testing.
For the fourth problem, we propose Credit-Aware Self-Critic (CASC) learning to fine-tune for matching the objective between training and testing. Different from previous methods that assign an identical reward for all actions, we reallocate the rewards based on the impacts of actor explorations.

The contributions are summarized as follows:
\begin{itemize}
    \item We leverage pre-trained language models to propose an extract-then-abstract framework, which contains a hierarchical extractor that efficiently handles long inputs while utilizing pretext knowledge.
    \item We investigate the problems for typical learning paradigms of the extractor and propose a framework, named \abbrev, to further improve the extractor performance. The studies on pseudo oracles also provide valuable insights for extract-then-abstract frameworks.
    \item Experimental results on Multi-News, Multi-XScience, and WikiCatSum corpora demonstrate that \abbrev~outperforms the state-of-the-art models with comparable sizes. 
\end{itemize}

\section{Related Work}

Early attempts for MDS focus on extracting sentences through statistical methods~\cite{10.3115/1117575.1117580,10.5555/1622487.1622501,wan-yang-2006-improved,10.1145/1390334.1390386}. For example, \citet{10.3115/1117575.1117580} extend Maximal Marginal Relevance (MMR) method to select sentences that are relevant to the query and novel across different documents. \citet{10.5555/1622487.1622501} leverage sentence relations with graph structures that represent pairwise sentence similarities, and apply PageRank~\cite{ilprints422} algorithm to extract sentences given the query document. However, extractive methods often suffer the coherence problem~\cite{Wu_Hu_2018}. Therefore, instead of directly extracting sentences from the articles, abstractive methods that can rewrite the articles achieve great success with the advantages of large annotated corpora~\cite{pang-etal-2021-agreesum,zhou-etal-2021-entity,liu-etal-2021-highlight,zhong-etal-2020-extractive,li-etal-2020-leveraging-graph,liu-lapata-2019-hierarchical}. 


Directly operating on the long inputs in MDS often leads to model degradation~\cite{jin-etal-2020-multi}. One of the promising solutions is to leverage the extract-then-abstract architectures in single document summarization~\cite{chen-bansal-2018-fast,gehrmann-etal-2018-bottom}. For example, \citet{gehrmann-etal-2018-bottom} propose an LSTM-based word-level extractor to choose phrases from the document, and apply an abstractor with the copy mechanism to generate summaries given the selected contents. \citet{pilault-etal-2020-extractive} also apply an LSTM-based sentence extractor to select contents, but further leverage a Transformer decoder to improve the performance. 

Although introducing Transformer architectures provides improvements, the input length is typically limited according to the computation and memory overhead. A recent line of studies has been proposed to alleviate such problems~\cite{brazinskas-etal-2021-learning,beltagy2020longformer,liu-lapata-2019-hierarchical,jin-etal-2020-multi}. \citet{liu-lapata-2019-hierarchical} propose to rank the candidate input paragraphs, and only concatenate the top few as the inputs for the abstractor. \citet{jin-etal-2020-multi} encode multiple documents in different granularity including token-level, sentence-level, and document-level. \citet{pasunuru-etal-2021-efficiently} design a graph encoder parallel with standard encoder to provide inter-document information and integrate the pre-trained BART~\cite{lewis-etal-2020-bart} with local attention mechanism~\cite{beltagy2020longformer} to overcome the long input problem. \citet{brazinskas-etal-2021-learning} apply policy gradient optimization to learn a train-time selector using lexical features pre-computed from source texts and gold summaries as inputs. Due to the lack of gold summaries in testing, a test-time selector is learned using lexical features solely from source texts as inputs and predictions from the train-time selector as learning targets. Comparatively, our methods thoroughly leverage the language models with an extract-and-abstract framework that enjoys the full capacities for both the extractor and the abstractor. The introduced summary referencing further improves the extractor with additional signals in both train- and test-phase.


\section{Methodology}
Utilizing large pre-trained language models brings great benefits for text generation problems. However, the input length of multi-document summarization is typically long, which makes large pre-trained language models, \textit{e.g.}, Transformer-based models, inefficient for processing. To match the length constraint, document-level truncation~\cite{fabbri-etal-2019-multi} has been widely-used. However, the truncation could inevitably cause information loss and degrade the performance. To solve the problem, we propose to leverage the extract-then-abstract framework. However, as described in the introduction, four challenges arise while learning such a framework. We first present our architecture to solve the long sequence problem and elaborate on the proposed methods to tackle the challenges.






\begin{figure*}[t]
\centering
\includegraphics[width=1.0\textwidth]{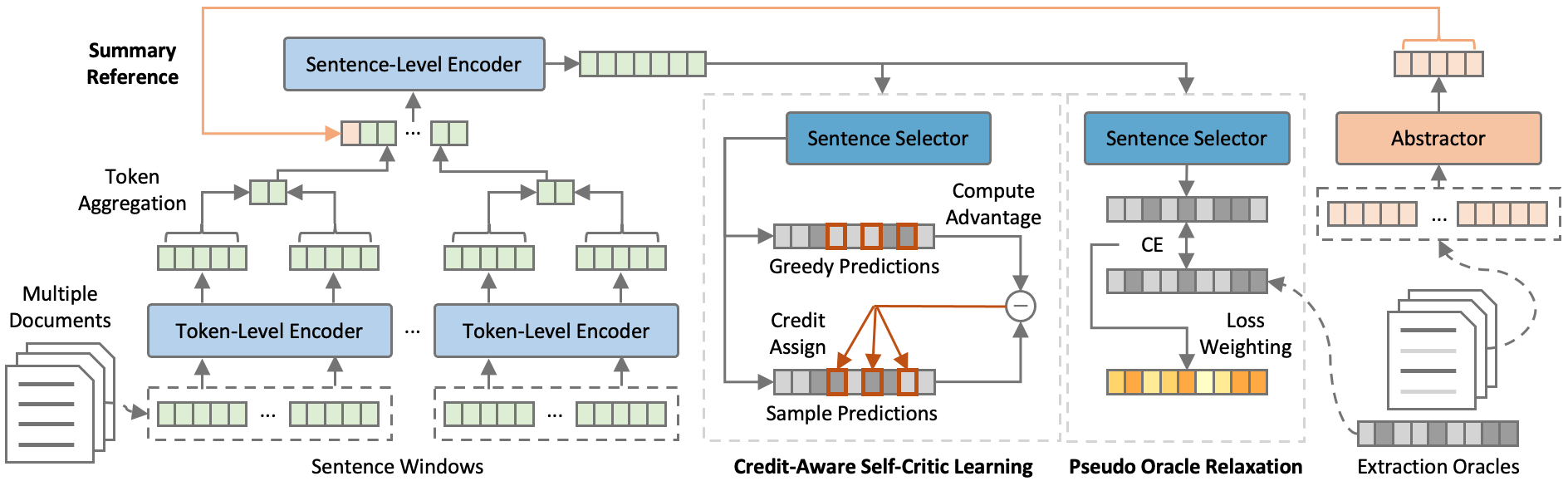}
\caption{Framework of~\abbrev. The illustration of proposed Pseudo Oracle Relaxation (POR), Summary Referencing (SR), and Credit-Aware Self-Critic (CASC) learning are indicated with bold fonts.}
\label{fg:framework}
\end{figure*}

\subsection{Hierarchical Summarizer}
Figure~\ref{fg:framework} illustrates the proposed REFLECT framework. Specifically, based on the Transformer~\cite{NIPS2017_3f5ee243}, our architecture contains a hierarchical extractor (H-EXT) and an abstractor (ABS). The hierarchical extractor is composed of a token-level encoder (T-ENC), a sentence-level encoder (S-ENC), and a sentence selector (SS). Considering a training example $(x,y)$, $x=\{x_1, x_2, ..., x_{N}\}$ is the concatenation of multiple documents that jointly consists of $N$ complete sentences and $y$ is the corresponding human-written summary. Let $M$ denote the allowed input length in Transformer-based models, we split all sentences into $K$ disjoint sets $\{h_k\}_{k=1}^K$ such that each set $h_k$ consists of sentences whose total number of tokens matches the length constraint $M$. The hierarchical extractor then predicts a set containing the indices of selected sentences (denoted by $\hat{e}$), which is used to retrieve salient contents from $x$ for the abstractor to produce summaries. Specifically, the hierarchical extractor can be expressed as follows.
\begin{equation}
\begin{split}
    \hat{e} & = \text{H-EXT}(h) \\
    & = \text{SS}(\text{S-ENC}(\bar{\oplus}\{\text{T-ENC}(h_k)\}_{k=1}^K)), 
\end{split}
\end{equation}
where $\bar{\oplus}$ is the operation of taking average for hidden states of tokens within sentences and concatenating the averaged results to obtain the sentence-level representations. The hierarchical summarizer (H-SUM) can then be expressed as:
\begin{equation}
    \text{H-SUM}(h) = \text{ABS}(\oplus\{ w_{x_i}|i \in \hat{e} \}),
\end{equation}
where $\oplus$ is the concatenation operation.

\subsection{Pseudo Oracle Relaxation}
\label{subsec:oracle_relax}
The sentence selection process of the extractor is typically formulated as a supervised classification problem. However, most summarization corpora do not provide such annotations. As an alternative, it is common to create pseudo oracles through a greedy process with the human-written summaries~\cite{liu-lapata-2019-text, chen-bansal-2018-fast}. However, such methods only provide suboptimal solutions since they are limited by the design of the greedy algorithm. The underlying patterns of true oracles could be too complicated to be designed precisely. The performance of the extractor is thus restricted since typical Maximal Likelihood Estimation (MLE) methods fully depend on pseudo oracles. For example, consider a case that there are only three sentences $x_a$, $x_b$ and $x_c$, and the ROUGE scores between sentences and summary are ranked as $x_a > x_b \gg x_c$. Assume that only $x_a$ is chosen as the pseudo oracle. If the extractor further selects one additional sentence for improving the results, it is expected that the combination of $(x_a, x_b)$ could be better than $(x_a, x_c)$ generally. However, MLE-based methods do not consider such discrepancy according to the pseudo oracle. 

Therefore, we propose a loss weighting mechanism to further consider the lexical similarity for the non-oracle sentences during the learning process. Consider an input $x=\{x_1, x_2, ..., x_N\}$ and the corresponding pseudo oracles $e$, the input sentences can be separated into two sets of pseudo oracles and non-pseudo oracles, $S = \{x_i | i \in e\}$ and $S^c=\{x_i | i \in e^c\}$, respectively. We maintain the loss weights for sentences in $S$ as one, while modifying the loss weights for the sentences in $S^c$ as follows:
\begin{equation}
    w_i = (1-\text{ROUGE-1}(x_i, y))^\gamma,
\end{equation}
where $\gamma$ is a hyper-parameter controlling the weighting scales. To stabilize the training, we further shift loss weights of $S^c$ such that the maximum value is one. This weighting mechanism emphasizes the predictions of both pseudo oracles and the low-ROUGE sentences, which have more impact on the performance. It relaxes the constraint from the binarized oracles to make the model aware of the differences between candidate sentences. The objective for the hierarchical extractor can then be written as:
\begin{equation}
    L = - \sum_{x_i \in x} w_i\log(\frac{exp(z^{\mathbf{1}_S(i)}_i)}{exp(z^0_i)+exp(z^1_i)}),
\end{equation}
where $z_i^0$ and $z_i^1$ are the binary logits for the $i$-th sentence, and $\mathbf{1}_S(i)$ is the indicator function of oracle sentences $S$.

\subsection{Summary Referencing}
\label{subsec:summary_referencing}
The pseudo oracle relaxation described in Section~\ref{subsec:oracle_relax} makes the extractor focus on the sentences that are more important in terms of the lexical similarity. In other words, it considers the discrepancy between non-oracle sentences instead of treating them equally. However, the ambiguity could still exist between the oracle and non-oracle sentences. For example, consider a case where there are two sentences $x_a$ and $x_b$, and only $x_a$ is chosen as the oracle during the greedy process. The ROUGE scores between the sentences and the summaries are roughly the same for the two sentences. Learning with such oracles may confuse the model since the positive and negative sentences are similar but with a large loss difference. Therefore, we propose to provide summary references for the extractor to further reason the selection. Since the summary is only available in the training stage, we leverage a fine-tuned abstractor to provide such summary references. With such a mechanism, the operations of the hierarchical extractor are further revised as:
\begin{equation}
\begin{split}
    r & = \text{ABS}(\oplus \{ w_{x_i}| i \in \hat{e} \}), \\
    \hat{e} & = \text{H-EXT}(h, r) \\
    & = \text{SC}(\text{S-ENC}(\bar{\oplus}( \{ \text{T-ENC}(r) \} + \\ 
    & \ \ \ \ \ \ \ \ \ \ \ \ \ \ \ \ \ \ \ \ \ \ \ \ \ \ \ \ \ \ \{\text{T-ENC}(h_k)\}_{k=1}^K))). \\
\end{split}
\end{equation}
Compared to directly using human-written summaries as references when training, which may cause the extractor highly rely on the reference signal, the generation results from the abstractor could serve as good approximations and provide the generalization from training to testing.

\subsection{Credit-Aware Self-Critic Learning}
With the objective of maximal likelihood estimation, the extractor is learned to select sentences that are jointly have high lexical similarity to the human-written summary. However, in extract-then-abstract frameworks, the required objective of extractor is to provide salient contents that can maximize the generation quality after rewritten by the abstractor. In other words, the objective for the extractor is inconsistent between training and testing. Therefore, we further propose a reinforcement learning method that can directly optimize with the test-time objective to bridge the gaps.

Specifically, we formulate the extraction of sentences as a single-round Combinatorial Multi-Armed Bandit (CMAB) problem~\cite{NIPS2016_aa169b49}. A general CMAB problem can be modeled as a tuple $(E,\mathcal{F},P,R)$, where $E=\{1,,2...,N\}$ is a set of $N$ arms, $\mathcal{F} \subseteq 2^E$ is a set of subsets of $E$, $P$ is a probability distribution over $[0,1]^N$, and $R$ is a reward function defined on $[0,1]^N \times \mathcal{F}$. At the $t$-th round, the agent pulls a subset of arms $S^t \in \mathcal{F}$, and produce stochastic outcomes $M^t=(M_1^t, M_2^t,...,M_N^t) \sim P$, where $M_i^t$ is the outcome of $i$-th arm. With a realization of outcomes $m=(m_1,m_2,...,m_N)$, the agents then receive a reward of $R(m,S)$. The goal of the agent is to maximize the expected cumulative reward in $T$ rounds, which is $\mathbb{E}_{\{M^t\}_{t=1}^T}[\sum_{t=1}^T R(M^t, S^t)]$. 

For the sentence extraction problem, we consider each sentence as an arm, and pull multiple arms (i.e., select multiple sentences) only for a single round. We solve this problem with the self-critic learning framework~\cite{Rennie_2017_CVPR}. Specifically, we consider the sentence-level encoder with the sentence selector that as the agent (jointly parameterized by $\theta$). The agent takes the sentence representations $x=\{x_1, x_2, ..., x_N\}$ from the token-level encoder, and selects a set of sentences $S$ according to a policy $\pi_\theta(\cdot)$ as:
\begin{equation}
\begin{gathered}
    m_i \sim\text{Bern}(\frac{exp(z^1_i)}{exp(z^0_i)+exp(z^1_i)}), \\
    S = \pi_\theta(x) = \{ i | m_i = 1 \},
\end{gathered}
\end{equation}
where $z_i^0$ and $z_i^1$ are the binary logits for the $i$-th sentence. To reduce the variance during learning, we introduce a baseline term through another policy $\tilde{\pi}_\theta$ as: 
\begin{equation}
\begin{gathered}
    \tilde{m}_i = \frac{exp(z^1_i)}{exp(z^0_i)+exp(z^1_i)}, \\
    \tilde{S} = \tilde{\pi}_\theta(x) = \{ i | \tilde{m}_i > 0.5 \},
\end{gathered}
\end{equation}
which is essentially a greedy policy as the taken actions are not explored. The two policies are derived from the agents with the same parameters, and the learning process is thus self-critical. We define the reward as the performance of generation from an abstractor using the selected sentences $S$ as the input, \textit{i.e.},
\begin{equation}
    R(S) = \text{ROUEG-L}(\text{ABS}(S), y),
\end{equation}
where $S$ is produced from the outcomes $m$. The advantage $a$ can be written as:
\begin{equation}
    a = R(S) - R(\tilde{S}).
\end{equation}
We then optimize the agent through gradient descent with the objective function as:
\begin{equation}
    L  = -\sum_{i \in E} a \log (\frac{exp(z^{\mathbf{1}_S(i)}_i)}{exp(z^0_i)+exp(z^1_i)}),
\end{equation}
where $\mathbf{1}_S(i)$ is the indicator function of selected sentence set $S$. However, in such a learning objective, the advantages are applied uniformly to update all actions, which could make learning difficult since the sentences number is typically large in our setting. Thus, we propose to specifically credit the advantages to the selections that are distinct between policies $\pi_\theta$ and $\tilde{\pi}_\theta$, and the objective function $L^{credit}$ can be written as:
    \begin{multline}
        L^{credit}  = \\
    -\sum_{i \in E} \mathbf{1}_{S \cap \tilde{S}}(i) a \log (\frac{exp(z^{\mathbf{1}_S(i)}_i)}{exp(z^0_i)+exp(z^1_i)}).
    \end{multline}

Different with previous work~\cite{chen-bansal-2018-fast} that formulates the sentence selections as a sequence of decisions and uses LSTM-based agent for learning, our formulation enables the usage of Transformer-based models for a better efficiency.

\begin{table*}[thb]
    \small
    \centering
    \begin{tabular}{lccccc}
        \toprule 
        Model & ROUGE-1 & ROUGE-2 & ROUGE-L &  Average & \makecell{Average \\ Improvement} \\
        \midrule
        Hierarchical-Transformer~\cite{liu-lapata-2019-hierarchical} & 42.36 & 15.27 & 22.08 & 26.57 & +17.91\% \\
        PG-BRNN~\cite{gehrmann-etal-2018-bottom} & 43.77 & 15.38 & 20.84 & 26.66 & +17.52\% \\
        Hi-MAP~\cite{fabbri-etal-2019-multi} & 44.17 & 16.05 & 21.38 & 27.20 & +15.18\% \\
        CTF-DPP~\cite{10.1613/jair.1.12522} & 45.84 & 15.94 & 21.02 & 27.60 & +13.51\% \\
        GraphSum~\cite{li-etal-2020-leveraging-graph} & 45.02 & 16.69 & 22.50 & 28.07 & +11.61\% \\
        GraphSum + RoBERTa~\cite{li-etal-2020-leveraging-graph} & 45.87 & 17.56 & 23.39 & 28.94 & +8.26\% \\
        Highlight-Transformer~\cite{liu-etal-2021-highlight} & 44.62 & 15.57 & - & - & - \\
        MatchSum ~\cite{zhong-etal-2020-extractive} & 46.20 & 16.51 & - & - & - \\
        PEGASUS~\cite{pmlr-v119-zhang20ae} & 47.52 & 18.72 & \textbf{24.91} & 30.38 & +3.13\% \\
        BART-Long ~\cite{pasunuru-etal-2021-efficiently} & 48.54 & 18.56 & 23.78 & 30.29 & +3.43\% \\
        BART-Long-Graph ~\cite{pasunuru-etal-2021-efficiently} & 49.24 & 18.99 & 23.97 & 30.73 & +1.95\% \\
        \midrule
        \abbrev~(MLE) & 48.16$\pm$0.01 & 18.87$\pm$0.01 & 23.78$\pm$0.17 & 30.27 & +3.50\% \\
        \abbrev~(CASC) & \textbf{49.27}$\pm$0.06 & \textbf{19.96}$\pm$0.03 & 24.76$\pm$0.09& \textbf{31.33} & - \\
        
        \bottomrule
    \end{tabular}
    \caption{Performance of \abbrev~with various baselines on Multi-News corpus. The results of Hierarchical Transformer, HiMAP, and PG-BRNN are copied from~\citet{li-etal-2020-leveraging-graph}. The rest baseline results are from the original papers. Best ROUGE scores are bolded. Performance of \abbrev~is reported with five runs.}
    \label{tab:main}
    \end{table*}

\section{Experimental Results}

\subsection{Settings}
\label{subsec:settings}
We compare \abbrev~with several strong baselines~\cite{liu-lapata-2019-hierarchical,gehrmann-etal-2018-bottom,fabbri-etal-2019-multi,10.1613/jair.1.12522,li-etal-2020-leveraging-graph,liu-etal-2021-highlight,zhong-etal-2020-extractive,pmlr-v119-zhang20ae,pasunuru-etal-2021-efficiently} on~\textbf{Multi-News}~\cite{fabbri-etal-2019-multi}, \textbf{Multi-XScience}~\cite{lu-etal-2020-multi-xscience} and \textbf{WikiCatSum}~\cite{perez2019generating} corpora, derived from news, academic domains and Wikipedia, respectively. Due to space limit, the results of Multi-XScience and WikiCatSum are provided in the Appendix~\ref{append:other_corpora}. For evaluation, we use ROUGE F1 metrics~\cite{lin-2004-rouge}, BERTScore~\cite{Zhang2020BERTScore:}, and factual consistency evaluated with FactCC~\cite{kryscinski-etal-2020-evaluating} to investigate performance from different perspectives. In our architecture, the hierarchical extractor is initialized by RoBERTa-base~\cite{liu2020roberta} containing 12 attention layers. We take the first $l$ layers and the rest layers (12-$l$) as the token- and sentence-level encoder, respectively. The input length limit $M$ is set to 512. The abstractor is a sequence-to-sequence model initialized by BART~\cite{lewis-etal-2020-bart}. To make the CASC computationally efficient, we use the BART-base as the abstractor that provides rewards for the extractor during training, while exploiting the BART-large in the testing. We provide more implementation details in Appendix~\ref{append:implementation_detail}.

\subsection{Main Results}
Table~\ref{tab:main} summarizes the performance of \abbrev~with various baselines on Multi-News corpus with the rows sorted by the average of ROUGE 1, 2, and L scores~\cite{lin-2004-rouge}. \footnote{We do not compare ROUGE-L results with Highlight-Transformer and MatchSum because 1) Highlight-Transformer does not report ROUGE-L results and 2) MatchSum reports summary-level ROUGE-L (ROUGE-LSum) results which are different from sentence-level ROUGE-L used here~\cite{lin-2004-rouge}. Nevertheless, ROUGE-LSum results of \abbrev~in Table~\ref{tab:ablation} still show the superiority over MatchSum (45.05 v.s. 41.89).} The results demonstrate that \abbrev~outperforms all baselines. PG-BRNN and Hi-MAP both use an LSTM-based point generator for sentence selection and apply a decoder for the generation. However, the performance is limited since LSTM could still suffer long-term dependency problem~\cite{trinh2018learning}. Hierarchical-Transformer (row 1) further uses an LSTM-based ranker to select paragraphs through predicting the ROUGE scores with the summaries for each paragraph, and uses a Transformer-based hierarchical encoder to capture the local and global information. However, the estimation of the ROUGE scores could be difficult since there are multiple valid summaries. In contrast, \abbrev~further applies pre-trained language models to extract the hierarchical features, and the proposed SR method explicitly provides the extractor with the evidence for selecting more salient contents to be rewritten from the abstractor. Moreover, GraphSum (row 5 \& 6) leverages graph structures to explore the relations of paragraphs in the encoder, and Highlight-Transformer (row 7) specifically assigns higher attention weights for key phrases. Both the methods implicitly provide additional selective information for the decoder. \abbrev~realizes such a content selection process in the data-level through the two-stage design. The proposed POR and SR methods alleviate the negative effects from the pseudo oracles, and thus provide complete information for the abstractor to rewrite. The results of PEGASUS (row 9) show the benefits of pre-trained sequence-to-sequence language models, which are also leveraged in \abbrev. The BART-Long-Graph (row 10 \& 11) further combines the previous advantages on graph structure and pre-trained language models to achieve strong performance. \abbrev~outperforms it in terms of the ROUGE-2 and ROUGE-L scores especially, probably due to the local attention mechanism used in BART-Long-Graph. Although such a design reduces the complexity to accommodate longer sequences, the trade-off for the attention capacity still restricts the model from generating more coherent summaries. Due to the space limit, please refer to Appendix~\ref{append:other_corpora} for results on Multi-XScience and WikiCatSum.

In addition to the evaluation of lexical overlaps, Table~\ref{tab:bs_factuality} shows the performance of semantic and factual consistency, which are important for summarization applications that will influence the public. The results demonstrate that \abbrev~can improve factual consistency through the hierarchical architecture, \textit{i.e.}, the architecture enables selection of useful information from multiple documents, while still maintaining the semantics in generations.



\begin{table}[h]
    \small
    \setlength\tabcolsep{5pt}
    \centering
    \begin{tabular}{lccc}
        \toprule 
        Model & BERTScore & Factual Consistency \\
        \midrule
        BART-base & 0.870 & 79.7 \\
        CTF-DPP & 0.852 & 81.9 \\
        \midrule
        \abbrev & \textbf{0.871} & \textbf{82.2} \\
        \bottomrule
    \end{tabular}
    \caption{BERTScore and factual consistency evaluated with FactCC on Multi-News corpus.}
    \label{tab:bs_factuality}
\end{table}

\begin{table*}[thb]
    \centering
    \small
    \begin{tabular}{cccccccccccc}
        \toprule 
        \multicolumn{2}{c}{MLE}
        & \multicolumn{2}{c}{RL}
        & \multicolumn{3}{c}{Extraction Performance}
        & \multicolumn{5}{c}{Abstraction Performance} \\
        \cmidrule(lr{0.5em}){1-2}\cmidrule(lr{0.5em}){3-4}\cmidrule(lr{1em}){5-7}\cmidrule(lr{1em}){8-12}
        SR & POR & SR & CA & Precision & Recall & F1
        & R-1 & R-2 & R-L & R-LSum & Average \\
        \midrule
        & & - & - & 0.6489 & 0.4218 & 0.5113 & 47.60 & 18.48 & 23.89 & 43.22 & 33.29 \\
        & \checkmark & - & - & 0.4618 & 0.7684 & 0.5769 & 47.92 & 18.76 & 23.57 & 43.87 & 33.53 \\
        \checkmark & & - & - & 0.4755 & 0.6275 & 0.5410 & \textbf{48.40} & \textbf{18.95} & \textbf{24.08} & 44.03 & \textbf{33.86} \\
        \checkmark & \checkmark & - & - & 0.4926 & 0.7112 & 0.5820 & 48.16 & 18.87 & 23.61 & \textbf{44.06} & 33.68 \\
        \midrule
        & \checkmark & & & 0.6161 & 0.4704 & 0.5335 & 48.59 & 19.04 & 24.20 & 44.21 & 34.01 \\
        & \checkmark & & \checkmark & 0.5344 & 0.5948 & 0.5630 & 48.91 & 19.77 & 24.46 & 44.80 & 34.49 \\
        \checkmark & & \checkmark && 0.5933 & 0.5328 & 0.5614 & 48.83 & 19.50 & 24.36 & 44.57 & 34.32 \\
        \checkmark & & \checkmark & \checkmark & 0.5442 & 0.5942 & 0.5681 & 49.04 & 19.80 & 24.50 & 44.94 & 34.57 \\
        \checkmark & \checkmark & \checkmark && 0.6007 & 0.5170 & 0.5557 & 48.84 & 19.41 & 24.34 & 44.56 & 34.29 \\
        \checkmark & \checkmark & \checkmark & \checkmark & 0.5873 & 0.5108 & 0.5464 & 
        \textbf{49.27} & \textbf{19.96} & \textbf{24.76} & \textbf{45.04} & \textbf{34.76} \\ 
        \bottomrule
    \end{tabular}
    \caption{Ablation of \abbrev~with Pseudo Oracle Relaxation (POR), Summary Referencing (SR), and Credit-Aware Self-Critic (CASC) learning. ROUGE score is abbreviated as \textit{R}.}
    \label{tab:ablation}
    \end{table*}

\subsection{Ablations and Analyses}
In this subsection, we perform ablations and analyses for \abbrev, and more results can be found in Appendix~\ref{append:additional_ablations}.

\noindent\textbf{Effect of Pseudo Oracle Relaxation (POR).} We start the discussions with the results using MLE objectives. As shown in Table~\ref{tab:ablation}, when applying POR to the base method (row 1 \& 2), the extraction recall increases while the precision decreases. Pseudo oracles generally contain sentences with high ROUGE scores to the summaries, and POR encourages the selection for such sentences by assigning smaller loss weights to increase the recall. From the abstraction performance, it shows that such selection preference could provide more concise information (even though not in pseudo oracles) as the input for improvements. Under the combination of SR (row 3 \& 4), the precision is further improved and achieves a higher overall F1 score due to more information for selection. However, the abstraction results show that the performance is actually inferior to the one only using SR, which meets our suggestion that the training objective of the extractor is inconsistent in the test-time. The proposed CASC learning further overcomes such problems and improves the performance by integrating POR and SR (row 6 \& 10).

\noindent\textbf{Effect of Summary Referencing (SR).} We first investigate the MLE results. With SR, the extractor is learned to select sentences given the approximations of summaries generated from the abstractor. The extraction recall is thus enhanced with such references. However, the trade-off between recall and precision still exists due to the discrepancy between ground-truth summaries and generation results (row 1 \& 3). Although ground-truth can be used as references to increase extraction performance during training, it fails to generalize in testing due to the distributional difference between them. The abstraction performance also shows SR makes significant improvements (row 3) and also benefits from the combination with CASC (row 6 \& 8).

\begin{figure}[h]
\centering
\includegraphics[width=0.5\textwidth]{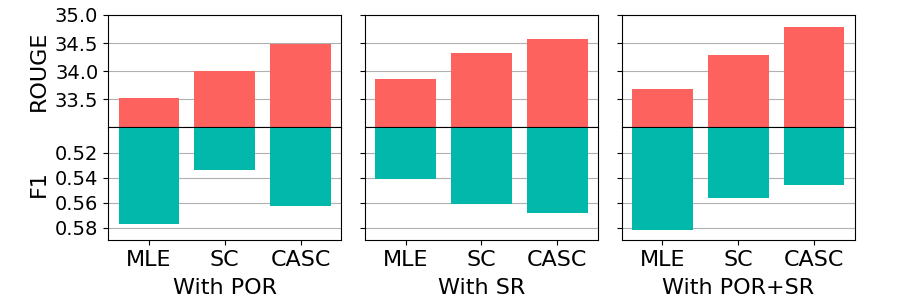}
\caption{Comparisons of MLE and RL results. For clarity, the displayed range for average ROUGE scores and extraction F1 scores are limited in [33,35] and [0.5,0.59], respectively.}
\label{fg:ablation}
\end{figure}

\noindent\textbf{Effect of Credit-Aware Self-Critic (CASC) Learning.} Consider the case that applying CASC with POR and SR respectively (row 5 \& row 6, row 7 \& row 8), the results demonstrate that CASC substantially outperforms the Self-Critic (SC) learning methods, even when combining the usage of POR and SR (row 9 \& row 10). We could further investigate the difference between MLE and RL methods through the visualization in Figure~\ref{fg:ablation}. It shows that, although having lower extraction performance, the RL methods can improve over the MLE methods in final abstraction performance due to the consistent objective between training and testing. CASC further improves over SC through explicitly assigning advantage to the actions that have the credits for exploration. Such design is critical for the long input sequences in multi-document summarization and can potentially be applied to other applications.

\subsection{Effects of Summary Reference}
\abbrev~uses the BART-large fine-tuned with truncated articles (BART-large-A) to provide summary references for both training and testing. Here, we study the effect of applying different summary referencing strategies in test-time. To investigate such effect across models, we leverage PEGASUS~\cite{pmlr-v119-zhang20ae} and fine-tune it with Multi-News corpus to produce summary references. Furthermore, we also conduct experiments that use the BART-large fine-tuned with pseudo oracle sentences (BART-large-O) and the ground-truth summaries. \textit{Note that these two methods are prohibited in practical usage since they require annotations on testing data.} The results in Table~\ref{tab:reference_study} manifest that using references from another model that is different from the training one does not significantly affect the performance. For the case using oracle inputs, the performance is similar to the ones directly using articles. This suggests that our method is stable for the practical scenario. The result of the ground-truth also shows that the reference quality could still affect the performance, which is left as a future research direction. 

\begin{table}[h]
    \small
    \setlength\tabcolsep{5pt}
    \centering
    \begin{tabular}{lccccc}
        \toprule 
        \makecell{Reference \\ Source} & R-1 & R-2 & R-L & Average \\
        \midrule
        PEGASUS & 49.28 & 19.99 & 24.84 & 34.78 \\
        BART-large-A & 49.29 & 19.99 & 24.83 & 34.79 \\
        \midrule
        BART-large-O & 49.29 & 19.98 & 24.81 & 34.78 \\
        Ground-Truth & 49.38 & 20.05 & 24.88 & 34.85 \\
        \bottomrule
    \end{tabular}
    \caption{Performance of different summary referencing strategies in \abbrev. We also present the results of BART models trained by truncated articles (A) and pseudo oracles (O). ROUGE score is abbreviated as \textit{R}.}
    \label{tab:reference_study}
\end{table}

\section{Conclusion}
In this work, we present an effective extract-then-abstract framework, named~\abbrev, for MDS. We utilize large pre-trained language models to construct a hierarchical extractor for solving the long sequence problem. Moreover, we investigate current learning paradigms for such frameworks and find that entirely relying on the pseudo oracles produced via a greedy process could hinder the performance. Therefore, we propose three corresponding techniques (POR, SR, and CASC) to overcome the issues. The experimental results not only show that~\abbrev~outperforms the state-of-the-art models on Multi-News, Multi-XScience, and WikiCatSum corpora, but also demonstrate that bridging the gap between training and testing is significant. Also, we present extensive studies to motivate more investigations. Finally, we consider further exploring the interactions between the extractor and abstractor, including iteratively providing a better reference or reusing the extraction predictions as the training signals for the abstractor, and study how to build a more efficient reference for providing the extraction evidence.

\section*{Acknowledgements}
This work is supported in part by the Ministry of Science and Technology (MOST) of Taiwan under the grants MOST-109-2221-E-009-114-MY3, MOST-110-2221-E-001-001 and MOST-110-2221-E-A49-164. This work was also supported by the Higher Education Sprout Project of the National Yang Ming Chiao Tung University and Ministry of Education (MOE), Taiwan. We are grateful to the National Center for High-performance Computing for computer time and facilities.



\bibliographystyle{acl_natbib}
\bibliography{anthology,custom}

\appendix
\section{Results on Other Corpora}
\label{append:other_corpora}
\subsection{\abbrev~on Multi-XScience Corpus}
In this subsection, we additionally experiment~\abbrev~with Multi-XScience~\cite{lu-etal-2020-multi-xscience} corpus, which is a multi-document summarization corpus for scientific articles. The source of this corpus is a tuple of a paper abstract and abstracts of the citation papers, while the generation target is the corresponding related-work paragraph. The performances of baselines are from ~\citet{lu-etal-2020-multi-xscience}~\footnote{We can not tell the version of score metric ROUGE-L used in Multi-XScience, and the evaluation code is also unavailable. Therefore, we do not report the ROUGE-L score from~\citet{lu-etal-2020-multi-xscience}}. Table~\ref{tab:xscience} shows that~\abbrev~achieves a better performance comparing to all baselines, and makes a substantial improvement especially in ROUGE-2 score. We also report the results of BART-large which serves as our abstractor to summarize from the extracted sentences. The improvement shows the benefit of the extraction process after the Credict-Aware Self-Critic Learning.

\begin{table}[h]
    \small
    \setlength\tabcolsep{4pt}
    \centering
    \begin{tabular}{cccccc}
        \toprule 
        \makecell{Model} & R-1 & R-2 & R-L & R-LSum \\
        \midrule
        LEAD & 27.46 & 4.57 & - & - \\
        LEXRANK & 30.19 & 5.53 & - & - \\
        TEXTRANK & 31.51 & 5.83 & - & - \\
        Hi-MAP & 31.66 & 5.91 & - & - \\
        PG & 34.11 & 6.76 & - & - \\
        \midrule
        BART-large & 33.29 & 8.07 & 17.31 & 29.04 \\
        \abbrev~(MLE) & 33.87 & 8.13 & 17.20 &  29.69 \\
        \abbrev~(CASC) & \textbf{34.18} & \textbf{8.20} & \textbf{17.42} & \textbf{29.73} \\
        \bottomrule
    \end{tabular}
    \caption{Performance of \abbrev~and baselines on Multi-XScience corpus. We also present the results of BART models trained by truncated articles. The results of baselines are from ~\citet{lu-etal-2020-multi-xscience}, where PG represents Pointer-Generator and the Hi-MAP represents the model proposed by~\citet{fabbri-etal-2019-multi}. ROUGE score is abbreviated as \textit{R}.}
    \label{tab:xscience}
\end{table}

\subsection{REFLECT on WikiCatSum Corpus}
To further analyze the capability of our proposed model, we also apply \abbrev~on a domain-specific corpus, WikiCatSum~\cite{perez2019generating} as shown in Table~\ref{tab:wikicatsum}. WikiCatSum is a multi-document summarization dataset derived from WikiSum~\cite{liu2018generating} and represents three distinct domains (Animal, Company, and Film). We compare our model with several baselines. TF-S2S is a Transformer sequence-to-sequence model of~\citet{liu2018generating} and CV-S2D+T is a variant of CV-S2S~\cite{gehring2017convolutional} with a single sequence encoder and a structure decoder. Both CV-S2D+T~\cite{perez2019generating} and TWAG~\cite{zhu2021twag} introduce the topic detection model to guide the generation. The decoder of CV-S2D+T is trained to predict the topics as an auxiliary task, while TWAG uses the topic information to group the input paragraphs and encodes them separately. ~\citet{liu2021noisy} apply knowledge distillation to alleviate the problem of single reference in maximum likelihood training, while our model leverages reinforcement learning to the train-test mismatch issue. The results of BART-large models are fine-tuned on WikiCatSum~\footnote{The models of ~\citet{liu2021noisy} and \abbrev~are with pretraining. We exploit the pretrained model from https://huggingface.co/facebook/bart-large-cnn.}. \abbrev~outperforms all baselines in all three domains, especially in the ROUGE-1 score, showing that our generated summaries carry more information. The results between \abbrev~(MLE) and \abbrev~(CASC) also manifest the effectiveness of Credit-Aware Self-Critic learning for bridging the gap between training and testing.

\begin{table}[thb]
    \small
    \centering
    \begin{tabular}{lcccc}
        \toprule
        \multicolumn{5}{c}{Animal} \\
        \midrule
        \makecell{Model} & R-1 & R-2 & R-L & R-LSum \\
        \midrule
        TF-S2S & 44.0 & 28.8 & 40.0 & - \\
        CV-S2D+T & 42.7 & 27.9 & 37.9 & - \\
        TWAG & 43.1 & 24.4 & 40.9 &  - \\
        \citet{liu2021noisy} & 45.9 & \textbf{32.2} & 41.4 & - \\
        \midrule
        BART-large & 46.3 & 29.2 & 39.6 & 44.1 \\
        \abbrev~(MLE) & 46.5 & 27.1 & 38.2 & 43.6 \\
        \abbrev~(CASC) & \textbf{48.6} & 30.2 & \textbf{41.5} & \textbf{46.0} \\
        \bottomrule
        
        \multicolumn{5}{c}{Company} \\
        \midrule
        \makecell{Model} & R-1 & R-2 & R-L & R-LSum \\
        \midrule
        TF-S2S & 26.0 & 9.5 & 20.4 & - \\
        CV-S2D+T & 27.5 & 10.6 & 21.4 & -\\
        TWAG & 34.1 & 11.9 & \textbf{31.6} & - \\
        \citet{liu2021noisy} & 33.5 & 15.0 & 25.9 & - \\
        \midrule
        BART-large & 36.8 & 15.1 & 25.6 & 33.6 \\
        \abbrev~(MLE) & 40.3 & 15.5 & 27.0 & 36.2 \\
        \abbrev~(CASC) & \textbf{40.8} & \textbf{15.8} & 27.5 & \textbf{36.6} \\
        \bottomrule
        
        \multicolumn{5}{c}{Film} \\
        \midrule
        \makecell{Model} & R-1 & R-2 & R-L & R-LSum \\
        \midrule
        TF-S2S & 36.5 & 18.8 & 31.0 & - \\
        CV-S2D+T & 38.0 & 21.2 & 32.3 & - \\
        TWAG & 40.8 & 21.2 & 34.3 & - \\
        \citet{liu2021noisy} & 42.7 & 26.1 & 36.8 & - \\
        \midrule
        BART-large & 44.3 & 25.5 & 35.9 & 41.7 \\
        \abbrev~(MLE) & 46.7 & 25.6 & 36.5 & 43.2 \\
        \abbrev~(CASC) & \textbf{47.6} & \textbf{26.8} & \textbf{37.9} & \textbf{44.1} \\
        \bottomrule
    \end{tabular}
    \caption{Performance of \abbrev~with various baselines on WikiCatSum corpus. The results of Transformer sequence-to-sequence (TF-S2S) and CV-S2D+T are referenced from ~\citet{perez2019generating}, and TWAG is from~\citet{zhu2021twag}. ~\abbrev~outperforms all baseline models in a large margin for all domains, especially for Company and Film in the ROUGE-1 score.}
    \label{tab:wikicatsum}
    \end{table}

\section{Additional Ablations \& Analyses}
\label{append:additional_ablations}
\subsection{Effects of Hierarchical Architecture}
In this subsection, we study the effect of hierarchical architecture in the extractor. As described in Section~\ref{subsec:settings}, the token- and sentence-level encoder jointly contain 12 layers in our settings. Thus, we experiment with different numbers of layers distributed for the token-level encoder. Also, we consider a case for no (0) layer, where we directly aggregate the word embeddings from the token-level encoder as sentence features. Table~\ref{tab:hierarchical_study} demonstrates that hierarchical architecture improves the resulted abstraction performance through better sentence representations. The performance is enhanced with the increasing number of layers, while more than 3 layers only makes marginal improvements. Thus, we use 3 token-level layers in this paper.

\begin{table}[h]
    \small
    \setlength\tabcolsep{5pt}
    \centering
    \begin{tabular}{cccccc}
        \toprule 
        \makecell{Layer \\ Number} & R-1 & R-2 & R-L & R-LSum & Average \\
        \midrule
        0 & 47.94 & 18.68 & 23.46 & 43.87 & 33.49 \\
        1 & 48.01 & 18.63 & 23.53 & 43.95 & 33.53 \\
        2 & 48.11 & \textbf{18.87} & \textbf{23.62} & 44.03 & 33.66 \\
        3 & \textbf{48.16} & \textbf{18.87} & 23.61 & \textbf{44.06} & \textbf{33.68}\\
        \bottomrule
    \end{tabular}
    \caption{Performance of different extractor configurations with MLE learning. Note that all configurations share the same amount of learnable parameters. ROUGE score is abbreviated as \textit{R}.}
    \label{tab:hierarchical_study}
\end{table}

\subsection{The Choice of Summary Reference at Training Stage}
In this subsection, we study the effect of the summery reference at training stage. As mention in the subsection~\ref{subsec:summary_referencing}, directly using human-written summaries as reference may cause the severe train-test mismatch. To verify the idea, we take the human-written summaries as the summery referencing for the extractor during training, and perform the standard test process, that is utilizing generated summaries as the references. The results shown in Table~\ref{tab:gt_summary_referencing} verify our assumption that training with such explicit signals reduces the ability of the model to generalize. Therefore, in our experiments, we take the generation results of BART-large as the summary referencing of the extractor.
\begin{table}[h]
    \small
    \setlength\tabcolsep{4pt}
    \centering
    \begin{tabular}{cccccc}
        \toprule 
        \makecell{MLE} & R-1 & R-2 & R-L & R-LSum & Average \\
        \midrule
        Ground-truth & 47.54 & 18.60 & 23.40 & 43.63 & 33.29 \\
        Generated  & 48.16 & 18.87 & 23.61 & 44.06 & 33.68\\
        \bottomrule
    \end{tabular}
    \caption{Performance of the MLE training results when taking the ground-truth summary as the summary reference (SR) during training. ROUGE score is abbreviated as \textit{R}.}
    \label{tab:gt_summary_referencing}
\end{table}

\subsection{Effect of Input Settings for Abstractor Fine-tuning}
In \abbrev, we decouple the learning for the extractor and the abstractor. Therefore, we study the effect of the input settings for the fine-tuning of the abstractor. Table~\ref{tab:bart_study} shows the performance of BART under different model configurations with either using truncated articles or pseudo oracle sentences as inputs. We find that using pseudo oracles for BART-large fine-tuning can effectively improve the performance even when the testing input is truncated articles. In addition, the results of using pseudo oracles for both training and testing demonstrate the upper bound performance of the extractor under MLE training, which suggests a promising development of our framework as the extractor can make more precise predictions.

\begin{table}[thb]
    \small
    \setlength\tabcolsep{5pt}
    \centering
    \begin{tabular}{cccccc}
        \toprule 
        BART & Train/Test & R-1 & R-2 & R-L &  R-LSum\\
        \midrule
        base & A/A & 45.71 & 17.12 & 23.82 & 41.54 \\
        base & O/A & 45.62 & 16.58 & 22.99 & 41.59 \\
        base & A/O & 49.53 & 22.02 & 26.78 & 45.01 \\
        base & O/O & 51.93 & 23.89 & 27.78 & 47.42 \\
        \midrule
        large & A/A & 46.80 & 18.01 & 23.80 & 42.57 \\
        large & O/A & 47.79 & 18.37 & 23.78 & 43.57 \\
        large & A/O & 51.01 & 22.77 & 27.00 & 46.53 \\
        large & O/O & 52.98 & 24.28 & 28.02 & 48.34 \\
        \bottomrule
    \end{tabular}
    \caption{Performance of BART with different configurations and train/test input settings. The input settings include truncated articles (A) and concatenated sentences with pseudo oracles (O). ROUGE score is abbreviated as \textit{R}.}
    \label{tab:bart_study}
    \end{table}

\subsection{Generation Examples}
We demonstrate two examples of generation in Table~\ref{tb:generation_example_1} and Table~\ref{tb:generation_example_2}. The results demonstrate that generations from \abbrev-(CASC) could provide more faithful information from multiple documents, which mainly resulted from the better sentence extraction strategy learned through CASC. 

\begin{table*}[t]
\renewcommand\arraystretch{1.0}
\centering
\small
\begin{tabular}{p{0.95\textwidth}}
\toprule
\textbf{Article:} Wesley Snipes Begins Serving 3-Year Prison Stint. This Wesley Snipes began serving a three-year sentence at a federal prison in Pennsylvania on Thursday for failure to file income tax returns. Snipes, 48, arrived shortly before noon at the Federal Correctional Institution McKean in the tiny northwestern Pennsylvania town of Lewis Run, federal prisons spokesman Ed Ross said. He had been ordered to surrender by noon. The minimum security prison camp is worlds away from the harsh prison fortresses depicted in the Snipes' films 'Undisputed' and 'Brooklyn's Finest.' The minimum-security camp doesn't have fences around its perimeter. The 300 nonviolent inmates live in barracks that feature two-man rooms, daily showers and double-feature movie showings Friday through Sunday. ...
\\
\midrule
\textbf{Gold Summary:} – Wesley Snipes enters prison in Pennsylvania today to serve a three-year sentence for tax evasion—and though the minimum-security McKean prison camp isn't exactly a five-star resort, it's not as bad as it could be. The AP gives details on what his life will be like: No fences: But he will submit to five daily head counts, three of which are overnight. Living quarters: He'll have a two-man room in the barracks, where the other 300 nonviolent inmates live. Daily schedule: Wake-up time is 6:35am, and jobs are performed for seven hours per day. Conjugal visits: He'll have to limit himself to just a kiss in the visitors room. Money: He can earn pennies an hour by doing laundry or other chores, and is allowed to spend \$290 a month at the commissary. Entertainment: Double-feature movies are shown Friday through Sunday, but no R-, NC-17-, or X-rated films are screened. Exercise: Sand volleyball, indoor basketball, exercise machines, badminton, bocci, and bridge will be available to him. Medical facilities: The copay at the infirmary is only \$2. Church: Almost any group you can think of, including Wiccans, hold weekly meetings. Click for more, including why Snipes failed to pay taxes—or watch his Tuesday night appearance on Larry King Live. \\
\midrule
\textbf{\abbrev~(MLE):} \textbf{– Wesley \textbf{\textbf{Snipes}}} has begun his \textbf{three\textbf{-year}} \textbf{\textbf{\textbf{prison}}} \textbf{sentence for} failure to file income \textbf{\textbf{tax}} returns. The Blade: Trinity \textbf{star} arrived at the Federal Correctional Institution \textbf{McKean} in Lewis Run \textbf{in Pennsylvania today}, reports \textbf{\textbf{the AP}}. The \textbf{minimum} \textbf{security} \textbf{prison camp is} worlds away from the harsh prison fortresses depicted in Snipes' \textbf{films} Undisputed and Brooklyn's Finest, notes the AP, but the 48-year-old actor will \textbf{have to} adjust to earning just \textbf{pennies an hour} handling kitchen, \textbf{laundry}, and other campus \textbf{chores}. "We recognize that he is high profile, but we treat all our \textbf{inmates} the same," says a spokeswoman. Snipes, a dues-paying member of a tax-protest \textbf{group} that challenges the government's right to collect \textbf{taxes}, has been free on bail since his 2008 conviction, but he made a last-minute request for a new trial this week, arguing that the judge erred by not allowing defense attorneys to interview jurors about misconduct allegations. He also told \textbf{Larry King} last night that he wasn't nervous about reporting to prison. "Don't send me up the river, yet," he said. \\
\midrule
\textbf{\abbrev~(CASC):} \textbf{– Wesley \textbf{\textbf{\textbf{\textbf{Snipes}}}}} has begun serving \textbf{a three-year sentence} in a \textbf{\textbf{minimum}-\textbf{security}} \textbf{\textbf{\textbf{prison}} in Pennsylvania} for failure to file income \textbf{tax} returns. The Blade: Trinity \textbf{star} arrived at the Federal Correctional Institution \textbf{McKean} yesterday, just before noon, reports \textbf{the AP}. He had been ordered to surrender by noon. The minimum security \textbf{prison camp is} worlds away from the harsh prison fortresses depicted in the Snipes' \textbf{films} Undisputed and Brooklyn's Finest. The \textbf{300 nonviolent \textbf{inmates} live} in \textbf{barracks} that \textbf{feature} \textbf{two-man} rooms, \textbf{daily} showers\textbf{, and} \textbf{double-feature} movie showings \textbf{Friday through Sunday}. The martial-arts enthusiast can get his \textbf{exercise} playing \textbf{sand volleyball} or \textbf{indoor basketball,} or work out on an elliptical machine or stair climber. Alas, no NC-17, R \textbf{or X} ratings \textbf{allowed}, which knocks out much of Snipes’ action-heavy repertoire. The most jarring aspect of the celebrity's stay might be the \textbf{five daily head counts, three} during the \textbf{overnight} \textbf{hours}. And Snipes, who earned a reported \textbf{\$}13 million for theBlade: Trinity sequel, will \textbf{have to} adjust to earning just \textbf{pennies an hour} handling kitchen, \textbf{laundry}\textbf{, or} other campus \textbf{chores}. \textbf{He can} \textbf{spend} just \textbf{\$290 a month at the} prison \textbf{commissary.} "We recognize that he is high profile, but we treat all our inmates the same," says a spokeswoman. Snipes made a last-minute request for a new trial on Wednesday, arguing that the judge erred by not allowing defense attorneys to interview jurors about misconduct allegations.
\\
\bottomrule
\end{tabular}
\caption{A generation example for \abbrev~on Multi-News corpus. Key overlaps between generations and the gold summary are bolded.}\smallskip
\label{tb:generation_example_1}
\end{table*}

\begin{table*}[t]
\renewcommand\arraystretch{1.0}
\centering
\small
\begin{tabular}{p{0.95\textwidth}}
\toprule
\textbf{Documents:} Vacant lot at 53 New York Avenue NE in Washington, where a government permit has been granted for work connected with Elon Musk’s Hyperloop project. (Michael Laris/TWP) It’s not much now, just a parking lot with a discarded gin bottle and an old exterminator receipt. But the slice of pavement near the Bureau of Alcohol, Tobacco, Firearms and Explosives in the District could be the gritty precursor to a tunnel network that could propel pods filled with people and speeding platforms topped with Teslas and Toyotas between the nation’s capital and New York in 29 minutes. Or it could be just be a parking lot littered with dashed transportation dreams. Electric-car pioneer and space entrepreneur Elon Musk has been touting his vision for a high-speed transportation system since his tweeted announcement last summer that he had received “verbal govt approval” for his tunnel-digging firm, the Boring Company, to build a “NY-Phil-Balt-DC Hyperloop.” The Boring Company team has received an early, and vague, building permit from the D.C. government...
\\
\midrule
\textbf{Gold Summary:} – – Elon Musk's dream of building a hyperloop that can move people between Washington, DC, and New York City in 29 minutes may be a small step closer to becoming a distant reality. A Nov. 29 permit issued by DC's Department of Transportation allows Musk's Boring Company to dig at an abandoned lot beside a McDonald's in northeast Washington that the company says might eventually become a station on a hyperloop connecting NYC, Philadelphia, Baltimore, and Washington. The system would speed up travel times, using electricity and electromagnetic levitation to send people and cars whipping through tunnels in pods. But before you ditch your car, the Washington Post points out the permit is only for "some preparatory and excavation work" at the site, and more permits are needed before any real construction begins. DC Mayor Muriel Bowser has visited a test tunnel Musk is constructing under Los Angeles. But though "we're open to the concept of moving people around the region more efficiently ... we're just beginning, in the mayor's office, our conversation to get an understanding of what the general vision is for Hyperloop," says Bowser's chief of staff, John Falcicchio. Still, the permit "is comparable to an athlete stretching before a race" and shows Musk is serious about moving forward with his plans for high-speed travel since claiming "verbal" government support for an East Coast hyperloop last July, reports TechCrunch. He's also received a conditional permit to build a tunnel beneath Maryland Route 295 in Hawthorne, per the Post, with Maryland officials saying a leg linking Washington and Baltimore will come first. (Read about the hyperloop pod's first test.) \\
\midrule
\textbf{\abbrev~(MLE):} \textbf{– Elon Musk's Boring Company} has \textbf{received a permit} from the \textbf{DC government} to begin \textbf{work on a} \textbf{tunnel}ing project in the nation's capital\textbf{, the Washington Post} \textbf{reports}. The permit, \textbf{issued} \textbf{Nov. 29}, is a "very \textbf{small}, very vague \textbf{step} \textbf{forward}" for Musk's plan \textbf{to build a} "NY-Phil-Balt-DC \textbf{Hyperloop}" tunnel network that would run from \textbf{\textbf{New York}} to \textbf{Philadelphia} to \textbf{Baltimore} to DC \textbf{in \textbf{29 minutes}}, according to the New York \textbf{Times}. But it could also be a "gravel precursor to \textbf{a tunnel} network" that could propel \textbf{pods} filled with \textbf{people and} speeding platforms topped with Teslas and Toyotas between the nation’s capital \textbf{and New York} in just 29 minutes. Musk's \textbf{company} has received backing from the White House \textbf{Office} of American Innovation, led by Jared Kushner, as well as \textbf{Maryland} Gov. Martin O'Malley and \textbf{DC Mayor Muriel \textbf{Bowser}}, who \textbf{visited a} tunnel in California earlier this month to learn more \textbf{about the} technology \textbf{\textbf{the company} says} will make tunneling faster and cheaper. The DC \textbf{Department of Transportation} is figuring out what other \textbf{permits} the company will need to cut under \textbf{city} roads and other public spaces. Asked whether the Bowser administration supports the project, he was somewhat upbeat but noncommittal, adding: \textbf{"We're open to the concept of moving people around the region more efficiently}."\textbf{
} \\
\midrule
\textbf{\abbrev~(CASC):} \textbf{– Elon \textbf{Musk}'s} plan \textbf{\textbf{to build} a} \textbf{high-speed} \textbf{\textbf{transportation}} \textbf{\textbf{system}} that would run from \textbf{New York} to \textbf{Philadelphia} to \textbf{\textbf{Baltimore}} to \textbf{\textbf{\textbf{DC}}} \textbf{in 29 \textbf{minutes}} has gotten a "very \textbf{small}, very vague \textbf{step} \textbf{forward}\textbf{\textbf{,"}} \textbf{the Washington Post} \textbf{reports}. \textbf{Musk's Boring Company} has \textbf{\textbf{received} a} \textbf{permit} for preparation and preliminary \textbf{excavation} of a \textbf{site} in \textbf{Washington, DC,} next to \textbf{a McDonald's} and near the Bureau of Alcohol, Tobacco, Firearms\textbf{, and} Explosives. It's not much now, just a parking \textbf{lot} with a discarded gin bottle and an old exterminator receipt, but it could be "the gritty precursor to \textbf{a \textbf{tunnel}} network that could propel \textbf{pods} filled with \textbf{people and} speeding platforms topped with Teslas and Toyotas between the nation’s capital \textbf{and New York} in \text{29 minutes}," \textbf{per \textbf{the Post}}. Or it could "be just be a Parking \textbf{Lot} littered with dashed transportation dreams." Musk has been touting his \textbf{vision} for \textbf{a \textbf{Hyperloop}} system since last summer, when he tweeted that he had received \textbf{"verbal} \textbf{government} approval" to build the "NY-Phil-Balt-DC Hyperloop." \textbf{Maryland officials} told the Post \textbf{that the} tunnel would run under \textbf{Maryland Route 295}\textbf{, with} the DC-Baltimore \textbf{leg} being built \textbf{first.} \textbf{DC Mayor Muriel} E. \textbf{\textbf{Bowser}} \textbf{visited} \textbf{\textbf{the company}} in California this month, walking in \textbf{a tunnel} to learn more \textbf{about the} technology \textbf{the company \textbf{\textbf{says}}} will make tunneling faster and cheaper. The District\textbf{'s Department of Transportation} is figuring out what other \textbf{permits} the company would need to cut under \textbf{city} roads and other public spaces, according to \textbf{Bowser's chief of staff}. \textbf{"\textbf{We}'re} \textbf{just beginning, in the mayor's office, our conversation to get an understanding of what the general vision is for Hyperloop,"} he says. Asked whether the Bowser administration supports the project, he says\textbf{, "}We’re \textbf{open to the concept of moving people around the region more efficiently}."\textbf{
}
\\
\bottomrule
\end{tabular}
\caption{A generation example for \abbrev~on Multi-News corpus. Key overlaps between generations and the gold summary are bolded.}\smallskip
\label{tb:generation_example_2}
\end{table*}
    
\section{Implementation Details}
\label{append:implementation_detail}
All of our experiments are conducted on a single NVIDIA
Tesla V100 32GB GPU with PyTorch. The hierarchical extractor is initialized by RoBERTa-base\footnote{https://huggingface.co/deepset/roberta-base-squad2}, and the first three layers are exploited as token-level encoder and the rest layers are sentence-level encoder. The BART-base\footnote{https://huggingface.co/facebook/bart-base} is used as the abstractor to provide the rewards during extractor training, while BART-large\footnote{https://huggingface.co/facebook/bart-large} is used for generating the final results. We generate the initial summary reference (SR) by the BART-large model. The hyper-parameter $\gamma$ in POR is set to 10. We use Adam with constant learning 3e-5 for optimization, and select the model with highest ROUGE-1 F1 score on validation set. For the pseudo extraction oracle, we greedily select at least 30 sentences from article as the pseudo oracle for the extractor during MLE learning. The selection criteria is based on the average of ROUGE-1 recall and ROUGE-2 recall.

For evaluations, we report ROUGE, BERTScore and factual consistency derived from FactCC framework. For ROUGE, we use \textit{rouge\_score} package \footnote{https://pypi.org/project/rouge-score} to report ROUGE-1, 2, L, and LSum scores. For BERTScore, we use official implementation \footnote{https://github.com/Tiiiger/bert\_score} and report the F1 scores. For factual consistency, we use official implementation of FactCC \footnote{https://github.com/salesforce/factCC} and follow previous work~\cite{10.1613/jair.1.12522} to calculate the factual consistency for multi-document summarization.

\end{document}